\documentclass[twoside]{article}

\usepackage[accepted]{aistats2026}
\usepackage[round]{natbib}

\newcommand\joonseok[1]{\textcolor{black}{#1}}

\newcommand\red[1]{\textcolor{red}{#1}}
\newcommand\green[1]{\textcolor{ForestGreen}{#1}}
\newcommand\greencell[1]{\cellcolor{green!10} \textbf{#1}}

\newcommand\graycell[1]{\cellcolor{gray!10} \textbf{#1}}

\usepackage{graphicx}
\usepackage{booktabs}
\usepackage{mathtools}
\usepackage{caption}
\usepackage{subcaption}
\usepackage{wrapfig}
\usepackage[most]{tcolorbox} 
\usepackage{multirow}
\usepackage{pifont}
\newcommand{\cmark}{\ding{51}}%
\newcommand{\xmark}{\ding{55}}%
\usepackage{enumitem}
\usepackage[dvipsnames,table]{xcolor}
\usepackage{pifont}

\usepackage[capitalise]{cleveref}
\crefname{table}{Tab.}{Tabs.}
\Crefname{table}{Tab.}{Tabs.}
\crefname{section}{Sec.}{Secs.}
\Crefname{section}{Sec.}{Secs.}
\Crefname{appendix}{App.}{Apps.}

\begin{document}

%

%
\runningauthor{C. Kim, S. Yi, Y. Kim, Y. Jo, J. Lee}

\twocolumn[
\aistatstitle{Towards Motion-aware Referring Image Segmentation}

\aistatsauthor{Chaeyun Kim$^{1,2*}$ \And Seunghoon Yi$^{1*}$ \And Yejin Kim$^{1}$ \And Yohan Jo$^{1}$ \And Joonseok Lee$^{1\dagger}$}
\vspace{0.5em}
\aistatsaddress{$^{1}$Seoul National University \qquad $^{2}$AIM Intelligence} ]

\renewcommand{\thefootnote}{\fnsymbol{footnote}}
\footnotetext[1]{Equal contribution.}
\footnotetext[2]{Correspond to: \texttt{joonseok@snu.ac.kr}}
\renewcommand{\thefootnote}{\arabic{footnote}}
\vspace{-0.5em}

\begin{abstract}
Referring Image Segmentation (RIS) requires identifying objects from images based on textual descriptions. We observe that existing methods significantly underperform on motion-related queries compared to appearance-based ones.
To address this, we first introduce an efficient data augmentation scheme that extracts motion-centric phrases from original captions, exposing models to more motion expressions without additional annotations.
Second, since the same object can be described differently depending on the context, we propose Multimodal Radial Contrastive Learning (MRaCL), performed on fused image-text embeddings rather than unimodal representations. For comprehensive evaluation, we introduce a new test split focusing on motion-centric queries, and introduce a new benchmark called M-Bench, where objects are distinguished primarily by actions. Extensive experiments show our method substantially improves performance on motion-centric queries across multiple RIS models, maintaining competitive results on appearance-based descriptions. Codes are available at https://github.com/snuviplab/MRaCL
\end{abstract}


\section{Introduction}
\label{sec:intro}

Referring Image Segmentation (RIS) is a task to identify and segment a target object described by a natural language expression from a given image.
This task is inherently multimodal,
requiring the model to associate visual and linguistic cues in a compositional (\textit{i.e.}, from combined linguistic elements) and contextual (\textit{i.e.}, based on its relation to the scene) manner.
With the advancement of vision-language models, the overall performance of RIS models has shown notable improvement~\citep{yang2022lavt,wang2022cris,hu2023beyond,ha2024nemo,yue2024adaptive}.


\begin{figure}[t]
    \centering
    \includegraphics[width=\linewidth]{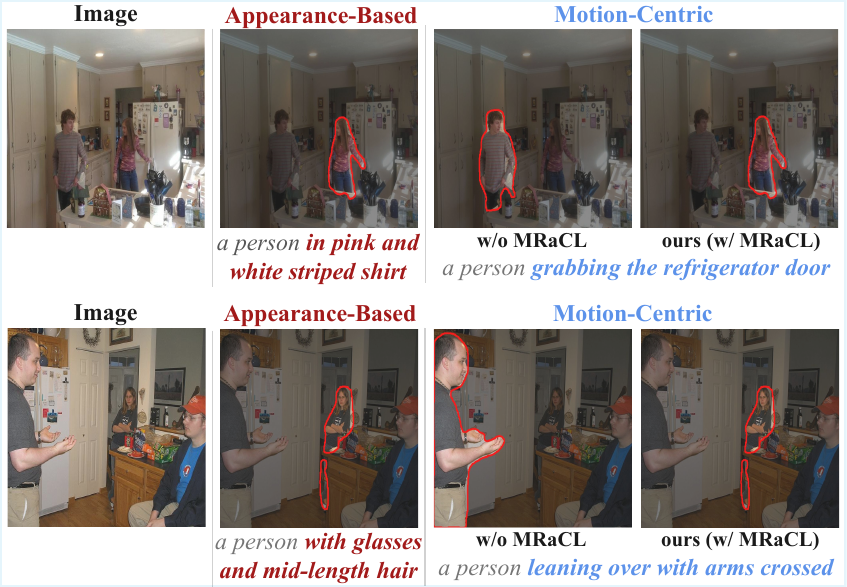}
    \vspace{-0.5cm}
    \caption{\textbf{Appearance-based (left) vs.\ motion-centric (right) queries in RIS.} Existing methods handle the former well but struggle with the latter.} 
    \label{fig:vis_static_motion_motivation}
    \vspace{-0.3cm}
\end{figure}

\begin{table}[t]
  \begin{minipage}{\columnwidth}
    \centering
    \footnotesize
    \renewcommand{\tabcolsep}{4pt}  
    \resizebox{\columnwidth}{!}{%
    \begin{tabular}{lcccc}
      \toprule
      & \multicolumn{2}{c}{\textbf{mIoU}} & \multicolumn{2}{c}{\textbf{oIoU}} \\ 
      \cmidrule(lr){2-3} \cmidrule(lr){4-5}
      \textbf{Models} & \textbf{Static} & \textbf{Motion} & \textbf{Static} & \textbf{Motion} \\ 
      \midrule
      CRIS~\citep{wang2022cris}  & 68.66 & 52.53 & 65.67 & 45.98 \\ 
      LAVT~\citep{yang2022lavt}  & 71.52 & 53.28 & 70.84 & 46.58 \\ 
      DMMI~\citep{hu2023beyond}  & 73.41 & 51.62 & 71.54 & 45.83 \\ 
      ASDA~\citep{yue2024adaptive} & 76.26 & 54.08 & 75.19 & 48.03 \\ 
      \bottomrule
    \end{tabular}}
    \vspace{-0.2cm}
    \caption{Performance gap between appearance-centric and motion-centric queries on G-Ref UMD test set.}
    \label{tab:intro_motion_stat}
    \vspace{-0.5cm}
  \end{minipage}
\end{table}


A key challenge in RIS stems from the complexity and variety of linguistic cues used 
in referring expressions.
In other words, the same object can be referred to in vastly different ways, including
visual appearance (\textit{red}, \textit{smiling}), multi-object relations (\textit{on the left, in front of}), and motions (\textit{walking, jumping}).
For instance, a person might be referred to as ``the person wearing an orange jacket'' (appearance-centric) or ``the person bending over and touching shoes'' (motion-centric).
Although both refer to the same target, the type of perception and reasoning required to distinguish this object from others can be significantly different.




Despite recent advances, we observe that even the state-of-the-art RIS models often struggle when motion-centric cues are given.
As illustrated in \cref{fig:vis_static_motion_motivation}, the model successfully segments the target when described with visual attributes in the second column, while fails when the same target is described by their motion, in the third column.
To verify whether this is a general trend, we manually curated two test sets, 300 appearance-centric and 300 motion-centric queries, from the G-Ref~\citep{mao2016generation} UMD split. 
\cref{tab:intro_motion_stat} confirms that all evaluated models show significant performance drops on motion-based queries, revealing an over-reliance on static visual attributes.



Why do the RIS models particularly suffer from motion-centric queries?
We hypothesize that 
insufficient exposure to diverse motion expressions during training is the primary cause.
Motion-related descriptions typically appear as verb phrases, and our analysis reveals that motion-oriented expressions have been underrepresented.
For instance, RefCOCO and RefCOCO+ contain only 10.4\% and 13.8\% of motion-centric queries, respectively.
G-Ref or Ref-ZOM contain more (40.3\% and 38.2\% each), but
such motion-oriented descriptions are often shadowed by easily identifiable accompanying nouns or subject descriptions.
This leaves motion-based expressions underutilized and hinders their ability to learn robust associations with the target.

To address this, we propose a data-augmentation approach to provide richer motion-centric training queries without requiring additional annotation.
Inspired by dropout \citep{srivastava2014dropout}, we propose to extract motion-based sub-part of the original textual description and utilize it as a positive during training, to encourage the model to learn diverse linguistic structures.
For instance, from a full expression ``a woman in orange shirt bending over'', we extract ``bending over'' as a supplementary positive description, guiding the model to localize targets with motion-oriented expressions without relying on other cues.


However, simply adding motion-centric expressions is insufficient.
In RIS, different expressions may refer to the same object only under specific image context.
For instance, ``a man with blue shirt'' and ``a man running on the street'' may refer to the same individual when conditioned on the image, but these convey different semantics in general.
Thus, simply placing embeddings of these two expressions closely is valid only under the given visual context; otherwise, such alignment may lead to incorrect associations.

Therefore, a naive use of the standard contrastive learning to unimodal representations would fail to capture this context-conditioned semantic overlap.
To this end, we propose to perform contrastive learning \emph{after fusing} the visual and textual embeddings.
Additionally, we observe that the conventional cosine similarity-based objective suffers from gradient saturation, hindering fine-grained training.
To resolve this problem, we design an alternative loss directly based on the angle between two embeddings, namely, Multimodal Radial Contrastive Loss (MRaCL).



Combining these ideas, our proposed approach augments with pairs of $\langle$image, motion-centric phrase$\rangle$, taking them as positives, and performs contrastive learning on their multimodal embeddings using our novel MRaCL loss.
It consistently improves performance across multiple baselines, particularly for motion-centric queries, while maintaining competitive accuracy on static expressions.
We further introduce M-Bench, a curated benchmark composed of action-centric RIS examples from video datasets, where the target object must be localized primarily via motion. 

We summarize our contributions as follows:
\begin{enumerate}
    \setlength{\itemsep}{0pt}
    \setlength{\parskip}{0pt}
  \item We identify the challenge of motion-centric queries in RIS and propose a novel textual augmentation strategy to mitigate underrepresented motion expressions.
  \item To better reflect the context-dependent nature of RIS, we propose MRaCL, a novel contrastive loss for image-conditioned semantic alignment of multimodal embeddings.
  \item We verify the effectiveness of our method across multiple RIS baselines and datasets, including our newly proposed motion-focused evaluation benchmark, M-Bench.
\end{enumerate}

\section{Related Work}
\label{sec:related}

\textbf{Referring Image Segmentation.}
The early architectures for reference image segmentation (RIS) extracted visual and linguistic features separately using CNNs~\citep{hu2016segmentation,liu2017recurrent} and RNNs~\citep{li2018referring,liu2017recurrent,margffoy2018dynamic,shi2018key}, which were then concatenated to form multimodal features. More recent work has shifted towards Transformer-based backbones~\citep{jing2021locate, liu2021swin, devlin2018bert, kamath2021mdetr} and multiscale features~\citep{chen2019see, hu2020bi, hui2020linguistic, ye2019cross} to capture finer object details and boundaries.

A central challenge in RIS is the effective fusion of multimodal features. Fusion strategies have progressed from simple concatenation~\citep{hu2016segmentation} to more sophisticated techniques. These can be broadly classified into three categories:
(1) \textit{Early fusion} methods such as LAVT~\citep{yang2022lavt} and DMMI~\citep{hu2023beyond} perform cross-modal interaction within the stages of the vision encoder.
(2) \textit{Late fusion} approaches including VLT~\citep{ding2021vision}, CRIS~\citep{wang2022cris}, and ReSTR~\citep{kim2022restr} encode modalities independently before merging them in a cross-modal decoder. 
(3) \textit{Multi-level fusion}, employed by CrossVLT~\citep{cho2023cross} and CoupAlign~\citep{zhang2022coupalign} introduces cross-modal interactions at multiple stages of the encoders.

\vspace{-0.1cm}
Recent works explore diverse set of architectural combinations. CGFormer~\citep{tang2023cgformer}, ASDA~\citep{yue2024adaptive}, and VATEX~\citep{nguyen2025vision} organize visual features into language-conditioned tokens to leverage global and local information. ReLA~\citep{liu2023gres} and DMMI~\citep{hu2023beyond} extend the task to support an arbitrary number of targets. Most recently, DETRIS~\citep{huang2025densely} has adopted parameter-efficient fine-tuning on pre-trained encoders with multiscale feature alignment.

\vspace{0.1cm} \noindent
\textbf{Deep Metric Learning.} 
Deep Metric Learning (DML) aims to learn embedding functions that map semantically similar instances closer together while pushing dissimilar ones apart in the feature space.
Foundational works such as contrastive loss~\citep{chopra2005learning}, $N$-pair loss~\citep{sohn2016improved}, and triplet margin loss~\citep{lee2018collaborative,lee2020large,ma2021contrastive} established early distance-based objectives using $L_p$ norms or dot-product similarity.

Subsequent approaches adopted softmax-inspired objectives to improve discriminability. For example, Center Loss~\citep{wen2016discriminative}, SphereFace~\citep{liu2017sphereface}, and CosFace~\citep{wang2018cosface} have been widely used in face recognition. Building upon this, ArcFace~\citep{deng2019arcface} introduced an angular-margin loss to enforce larger angular separations. Recently, AoE~\citep{li2024aoe} applied these ideas to semantic textual similarity tasks while addressing gradient saturation issue common in cosine-based losses.

Although the principles of DML have been increasingly applied to multimodal domains, a key limitation remains: standard contrastive objectives, often relying on cosine similarity, suffer from gradient saturation and is thus suboptimal
in the inherently anisotropic spaces of pre-trained encoders~\citep{gao2021simcse,li2024aoe}.
Our proposed Multimodal Radial Contrastive Loss (MRaCL) addresses this limitation by employing an angle-based similarity metric within a contrastive framework on fused multimodal embeddings, enabling more effective fine-grained alignment for the RIS task, as detailed in \cref{sec:method:mracl_loss}.

\begin{figure*}[t]
    \vspace{-0.2cm}
    \centering
    \includegraphics[width=\linewidth]{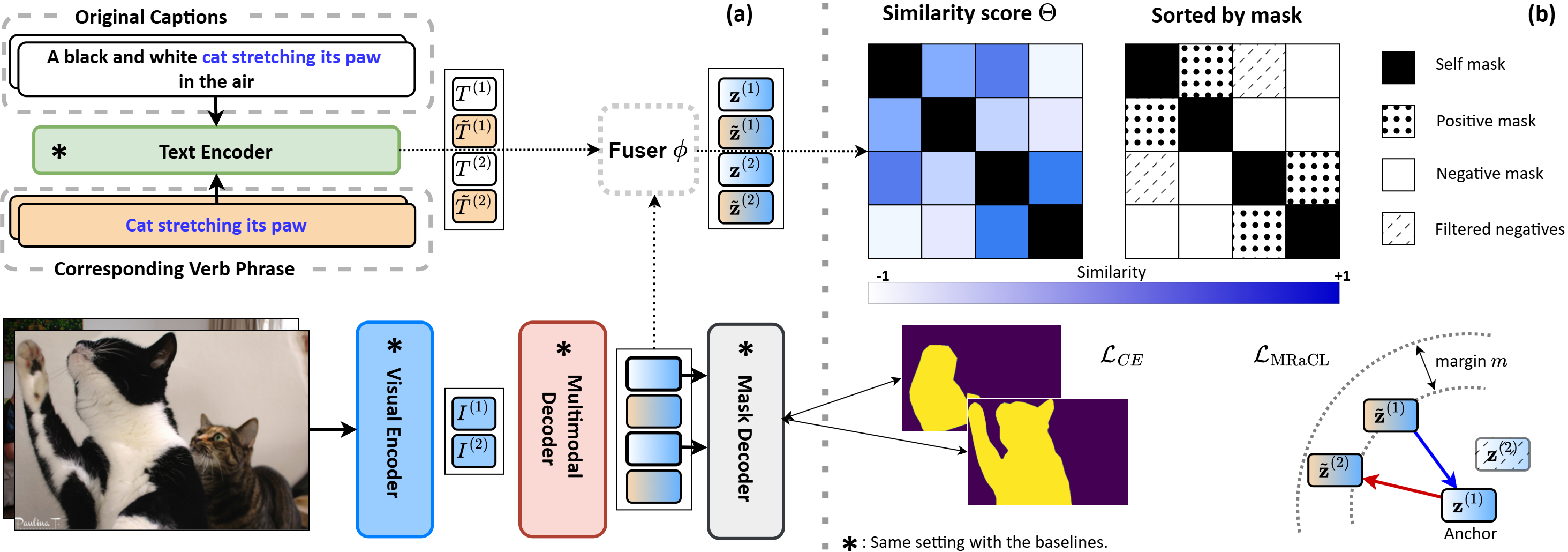}
    \vspace{-0.6cm}
    \caption{
    \textbf{Overview of the MRaCL framework}, (a) We apply CE Loss to output masks of the original pairs, same as the baselines. The \textit{Fuser} mix text embeddings and cross-modal embeddings from the decoder, returns multimodal representations $\textbf{z}^{(i)}$. (b) With those multimodal latents, we calculate similarity scores and filter out false negatives. For example, with $\textbf{z}^{(1)}$ as the anchor, $\textbf{z}^{(2)}$ is considered as false negative sample and masked out as shown in the bottom right diagram. Here, components with marked as `\textbf{$*$}' indicates that it originated form the baseline model. }
    \label{fig:mracl_architecture}

    \vspace{-0.3cm}
\end{figure*}

\section{Method}
\label{sec:method}

Our approach enhances RIS models' understanding of action-centric expressions through two complementary strategies. First, we propose a data augmentation strategy that extracts action-focused phrases from referring expressions. Second, recognizing that semantically different phrases can refer to the same object only within a specific image context, we introduce Multimodal Radial Contrastive Loss (MRaCL) that performs contrastive learning on fused image-text embeddings rather than unimodal representations.
Observing that simply exposing the model to more verb phrases is not sufficient, however, we further propose a more direct enhancement in RIS modeling and training.
Considering the unique characteristics of RIS that semantically different phrases can refer to the same object conditioned on an image, it is often misleading to simply align textual embeddings referring to the same object.
Instead, we propose to align them in conjunction with the conditioned image, so that the model can learn semantic proximity between multimodal embeddings, rather than unimodal textual embeddings.
Furthermore, we introduce a novel objective called Multimodal Radial Contrastive Loss (MRaCL), designed to expedite the fine-grained alignment
in a shared multimodal embedding space.
Together, these strategies enable the model to capture nuanced action semantics and foster more robust compositional understanding in RIS.
We describe each idea in detail in this section.



\subsection{Data Augmentation with Motion-centric Verb Phrase Retrieval}
\label{sec:method:hpconstruct}

Existing RIS models often overlook action-centric expressions, as objects can be often distinguished by appearance alone.
We hypothesize that exposing a model to more training examples where the target is disambiguated only through its motion will improve the model's motion understanding capabilities.
However, this approach can introduce ambiguity when multiple objects of the same category perform similar actions (\textit{e.g.}, two people `jumping over'). To mitigate this potential issue, we apply a pre-filtering strategy to our training data, retaining only those samples where the target's category appears a single time. The effectiveness of this strategy is validated in \cref{tab:ambiguity_filtering}.
To this end, we propose to augment the training set by adding \emph{verb phrases} which highlights the motion extracted from the original expression.

More formally, given a training example $(I, T)$ where $I$ is an image and $T$ its textual description, we extract verb phrases from $T$ that capture the referent's motion, using an LLM (\emph{e.g.}, LLaMA 3.1-70B~\citep{grattafiori2024llama}) with a customized prompt that guides it to \emph{retrieve} motion-oriented verb phrases from the original caption, ensuring that they are grounded in the input description.
The LLM identifies whether $T$ contains such verb phrase and, if so, extracts the most detailed and contextually relevant span, denoted by $\tilde{T}$.
The resulting $\tilde{T}$ may include a single verb (`running'), a verb-object pair (`doing tricks'), or a longer construct with prepositional phrases (`placed on the table'), covering both active motions and passive states.

Crucially, we utilize $(I, \tilde{T})$ as a \emph{supplementary}, not as a replacement for the training example, alongside the original $(I, T)$, 
 thereby \textit{retaining original cues} 
from $T$ while reinforcing the model's attention to motion-centric semantics through $\tilde{T}$. 
Training with the supplementary examples, the model is guided to learn that both $T$ and $\tilde{T}$ refer to the same target.
By incorporating these partial, motion-centric expressions into training, we encourage the model to correctly localize the target even in the absence of static attributes or complete contextual details.







\subsection{Multi-modal Contrastive Learning}
\label{sec:method:MCL}
We empirically observe, however, that simply providing additional training examples does not help.
In fact, naively performing contrastive learning on these augmented data actually drops the RIS performance.
This stems from a unique characteristic of RIS: an object can be described with various aspects (\emph{e.g.}, appearance, motion, relation with others), and these expressions are semantically aligned only when they are conditioned on the image.
For example, `a lady wearing yellow shirt' and `a lady running on the street' convey completely uncorrelated semantics in general, but they may refer to the same object conditioned on a particular image.
For this reason, simply learning to align them without grounding in a visual context would confuse the model.
We resolve this challenge by implementing a cross-modal contrastive learning framework, which we briefly review first and develop our ideas.


\vspace{0.1cm} \noindent
\textbf{Review of Contrastive Learning.} 
As a representative metric learning method, contrastive learning has become a cornerstone for representation learning.
SimCSE~\citep{gao2021simcse} introduced a effective way to learn sentence representations, subsequently adopted by many variants~\citep{zhou2022debiased, zhang2022unsupervised, chuang2022diffcse, zhang2022contrastive, liu2023rankcse}.
Given a batch $\mathcal{B} = \{\mathbf{z}_i\}_{i=1}^n$ of text embeddings, SimCSE creates positive pairs by encoding each caption twice with dropout-based augmentation.
Taking the other augmented one from the same text as a positive and all other pairs in the batch as negatives~\citep{chen2017sampling}, SimCSE minimizes
the NT-Xent loss~\citep{chen2020simple}
\begin{equation}
    \mathcal{L}_{\text{NT-Xent}} = -\log \frac{e^{\text{sim}(\mathbf{z}^{(i)}, \mathbf{\tilde{z}}^{(i)}) / \tau}}{\sum_{j=1}^n e^{\text{sim}(\mathbf{z}^{(i)}, \mathbf{z}^{(j)}) / \tau}},
    \label{eq:xent_loss}
\end{equation}
where $\mathbf{z}^{(i)}$ is the anchor, $\mathbf{\tilde{z}}^{(i)}$ is the positive sample of $\mathbf{z}^{(i)}$, and $\mathbf{z}^{(j)}$ ($j \neq i$) are negative samples, and $\tau$ is the temperature hyperparameter.
The sim($\mathbf{x}, \mathbf{y}$) is a similarity function between $\mathbf{x}$ and $\mathbf{y}$, \emph{e.g.}, the cosine similarity $\text{sim}(\mathbf{x}, \mathbf{y}) = \mathbf{x}^\top \mathbf{y} / \|\mathbf{x}\| \|\mathbf{y}\|$.


\vspace{0.1cm} \noindent
\textbf{Extension to Multimodal Embedding Space.}
To ensure that semantically different texts referring to the same object are aligned only within their specific visual context, we propose to conduct contrastive learning on \textit{fused cross-modal representations} rather than unimodal representations in RIS.



Given an image-text pair $(I, T)$, we extract a cross-modal embedding $\mathbf{z} = \phi(\mathbf{x}_I, \mathbf{x}_T)$, where $\mathbf{x}_I$ and $\mathbf{x}_T$ indicate the image and text features from corresponding pre-trained models (see \cref{tab:refcoco_results} for details).
$\phi$ is a visual-text fuser, 
which we use a single projection layer, in \cref{fig:mracl_architecture}(a).
$\mathbf{z}$ serves as our anchor, as shown in the lower right of \cref{fig:mracl_architecture}(b).
Similarly, we also extract the cross-modal embedding from the augmented pair $(I, \tilde{T})$, yielding a positive pair embedding $\tilde{\mathbf{z}} = \phi(\mathbf{x}_I, \mathbf{x}_{\tilde{T}})$.
Other pair embeddings such as $\phi(\mathbf{x}_I^{(i)}, \mathbf{x}_T^{(j)})$ for $j \ne i$ are considered as \textit{negatives}, where $i, j$ are indices within a minibatch.
With this redefined anchor, positive and negatives, we can apply contrastive learning, \emph{e.g.}, via minimizing the NT-Xent loss in Eq.~\eqref{eq:xent_loss} or its variants, in the multimodal space.

\vspace{0.1cm} \noindent
\textbf{False Negative Elimination.}
In the RIS setting, where image–text pairs are usually not in a one-to-one correspondence, semantically similar descriptions often appear across different samples.
For instance, ``person walking'' and ``moving on the street'' describe similar actions, and they may refer to the similar target from different images.
Thus, treating them as a negative may not be desirable.

Inspired by the false negative cancellation strategy~\citep{huynh2022boosting}, we design an optional filtering step to identify and filter out potential false negatives leveraging the model’s own multimodal embeddings.
Specifically, we first calculate similarity score $\Theta$
of anchor embeddings within each mini-batch, as in \cref{fig:mracl_architecture}(b).
When the score between an anchor $\mathbf{z}^{(i)}$ and a nominal negative exceeds a threshold $\nu$ (empirically 0.5), we consider that pair as a potential \textit{false negative} and exclude from the contrastive loss computation.

\subsection{Multimodal Radial Contrastive Learning}
\label{sec:method:mracl_loss}
\begin{figure}[t]
    \centering
    \includegraphics[width=\linewidth]{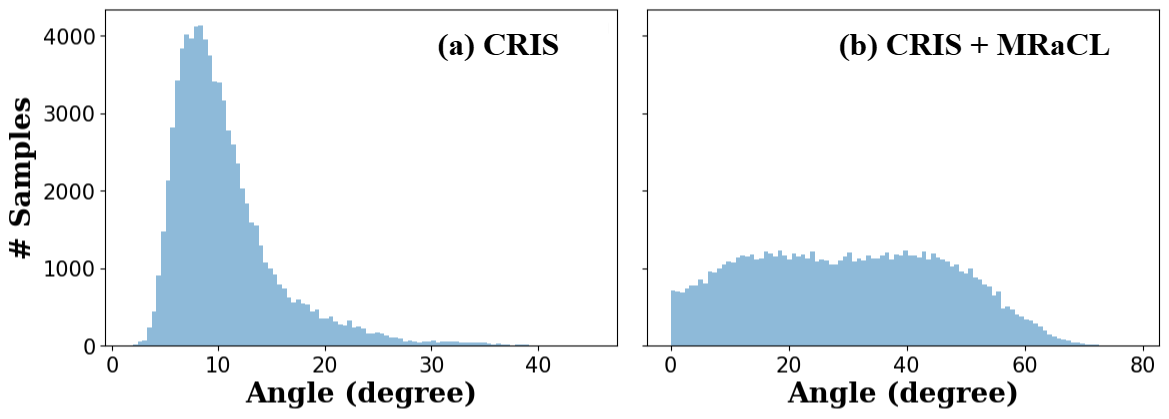}
    \vspace{-0.7cm}
    \caption{\textbf{Anisotropy phenomenon observed in a baseline model (CRIS)}. We plot the distribution of pairwise angular distances of 100K pairs, (a) from the original CRIS and (b) with our MRaCL loss. We clearly see that the model utilizes a significantly wider representation space using our MRaCL loss.
    }
    \label{fig:angular_loss_reason}
    \vspace{-0.3cm}
\end{figure}

Although we have addressed the issue with action-centric queries and contrastive learning from the RIS perspective, the contrastive objective based on cosine similarity remains suboptimal, for two reasons.
First, the cosine similarity suffers from \emph{similarity saturation}~\citep{li2024aoe}.
As the similarity approaches $\pm$1, its gradient significantly diminishes, providing negligible update signals when embeddings are already close (positive pairs) or distant (negative pairs).
This hinders learning of fine-grained distinctions crucial for differentiating subtle action variations.
Second, multimodal embeddings exhibit \emph{anisotropy}, which tend to cluster in confined regions of the embedding space.
Primarily observed in pre-trained language models~\citep{gao2019representation, ethayarajh2019contextual, wang2019improving, li2020sentence}, it also occurs in RIS models. 
For example, empirical pairwise angular distances between embeddings from CRIS (in \cref{fig:angular_loss_reason}(a)), reveals that most embeddings are tightly clustered within a small angular distance. 
This exacerbates the saturation problem, making it particularly challenging to differentiate fine-grained action expressions.



To address these issue, we adopt the \emph{angular distance} as our similarity metric, building upon AoE~\citep{li2024aoe}. 
Specifically, for two normalized multimodal representations, $\mathbf{z}^{(i)}$ and $\mathbf{z}^{(j)}$, we define our similarity metric $\theta_{i,j}$ as
\begin{equation}
\footnotesize
    \theta_{i,j} = \tfrac{\pi}{2} - \angle (\mathbf{z}^{(i)}, \mathbf{z}^{(j)}) = \tfrac{\pi}{2} - \cos^{-1}\bigl(\text{sim}(\mathbf{z}^{(i)}, \mathbf{z}^{(j)})\bigr),
    \label{eq:theta}
\end{equation}
where $\angle(\mathbf{x}, \mathbf{y})$ indicates the angle between the two vectors $\mathbf{x}$ and $\mathbf{y}$, and
$\text{sim}(\cdot, \cdot)$ denotes cosine similarity.
We subtract from $\pi/2$ to convert the angle to a similarity score.
This choice mitigates the similarity saturation problem, as the derivative $d \theta_{i,j} / d\mathbf{z}$ does not vanish unless $\theta_{i,j} = \pi/2$, thereby maintaining 
substantial gradients even for closely aligned embeddings.

For a more robust alignment, we introduce a margin $m \ge 0$ on the positive pair similarities $\theta_{i, \tilde{i}}$.
This enforces a minimum angular separation between positive and negative pairs, encouraging the model to learn more discriminative representations.
Also, it mitigates anisotropy by explicitly penalizing embeddings that are too close, driving them to utilize a broader embedding space.


Replacing the similarity score with Eq.~\eqref{eq:theta} and combining all the elements, we get our Multimodal Radial Contrastive Loss (MRaCL): 
\begin{equation}
\footnotesize
    \mathcal{L}_{\text{MRaCL}} 
    = -\log \frac{\exp\bigl((\theta_{i,\tilde{i}}-m)/\tau\bigr)}
    {\exp\bigl((\theta_{i,\tilde{i}}-m)/\tau\bigr) + \sum_{j \ne i} \exp(\theta_{i,j}/\tau)},
    \label{eq:final_loss}
\end{equation}
where $\theta_{i,\tilde{i}} = \pi/2 - \angle(\mathbf{z}^{(i)}, \mathbf{\tilde{z}}^{(i)})$ following Eq.~(\ref{eq:theta}).

Finally, integrating with the conventional segmentation loss $\mathcal{L}_\text{seg}$ (\emph{e.g.}, pixel-level mask cross-entropy) completes our full training objective:
$\mathcal{L} = \mathcal{L}_{\text{seg}} + \alpha \cdot \mathcal{L}_{\text{MRaCL}}$, with 
a balancing weight $\alpha$.

\begin{figure*}[t]
    \centering
    \includegraphics[width=\linewidth]{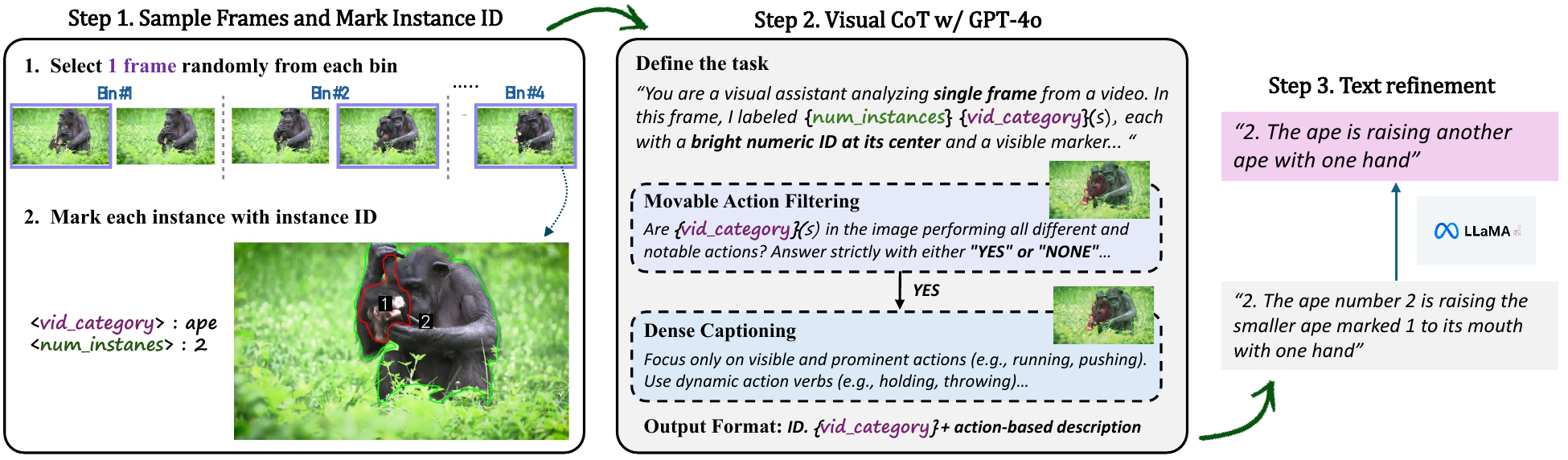}
    \caption{\textbf{Overview of the MBench dataset annotation pipeline.}}
    \label{fig:mbench_pipeline}
\end{figure*}

\section{Experiments}
\label{sec:exp}
\vspace{-0.4cm} \noindent

\subsection{Experimental Setups}
\label{sec:exp:setting}
\vspace{-0.2cm} \noindent

\textbf{Datasets.} We evaluate our method on four widely used RIS benchmarks: RefCOCO~\citep{yu2016modeling}, RefCOCO+~\citep{kazemzadeh2014referitgame}, G-Ref~\citep{mao2016generation} and Ref-ZOM~\citep{hu2023beyond}.
RefCOCO contains positional phrases like ``front'' or ``third from the right,'' while RefCOCO+ excludes such cues.
G-Ref features longer and more descriptive expressions (9--10 words on average), and Ref-ZOM allows cases with no target or multiple targets.
Among these, G-Ref and Ref-ZOM include more motion-centric cues and object relationships, making them particularly suitable for evaluating our approach.


\noindent
\textbf{Motion-Oriented Benchmarks.} 
\label{sec:exp:setting:mbench}
To systematically evaluate motion understanding in RIS, we introduce two complementary benchmarks. First, \textbf{M-Ref} consists of 325 appearance-centric and 325 motion-centric queries manually sampled from the G-Ref (UMD) test set, enabling controlled comparison between query types (examples in App.~\ref{appendix:mref}).

Second, we construct \textbf{M-Bench}, a comprehensive motion-centric benchmark built upon three video object segmentation datasets: A2D-Sentences~\citep{gavrilyuk2018actor}, Ref-Youtube-VOS~\citep{seo2020urvos}, and ViCaS~\citep{athar2024vicas}. Leveraging their pixel-level mask annotations, we generate fine-grained action-oriented captions aligned with each object through a two-stage pipeline (\cref{fig:mbench_pipeline}).

We initially sample representative frames and retain only those containing at least two instances of the same movable category (\textit{e.g.}, multiple persons), ensuring that targets cannot be identified by category alone.
Then, we employ visual chain-of-thought prompting with an MLLM to produce a concise, motion-centric description for each instance, followed by refinement with a lightweight language model.
Every description is manually verified, and we discard samples whose captions do not reflect the actual behavior or where multiple instances exhibit indistinguishable actions. 
This yields 8,200 evaluation-only samples designed to assess action-based expression understanding.
Full details are in ~\cref{appendix:mbench,appendix:mbench_construction}.

\begin{table*}[t]
\centering
\label{tab:refcoco_results}
\setlength{\tabcolsep}{1pt}
\resizebox{\linewidth}{!}{
\begin{tabular}{lc||rrr|rrr|rr|c||rr|rr|rr}\toprule

\multirow{3}{*}{\textbf{Method}} & \multirow{3}{*}{\textbf{+MRaCL}} & \multicolumn{3}{c|}{\textbf{RefCOCO(UNC)}} &\multicolumn{3}{c|}{\textbf{RefCOCO+(UNC+)}} &\multicolumn{2}{c|}{\textbf{G-Ref(UMD)}} &\textbf{Ref-ZOM} &\multicolumn{4}{c|}{\textbf{M-Ref 
}} &\multicolumn{2}{c}{\textbf{MBench}} \\ 

&& & \multicolumn{1}{c}{oIoU} & & & \multicolumn{1}{c}{oIoU} & & \multicolumn{2}{c|}{oIoU} & \multicolumn{1}{c||}{oIoU} &\multicolumn{2}{c|}{mIoU} &\multicolumn{2}{c|}{oIoU} &mIoU &oIoU \\ 
& & \multicolumn{1}{c}{val} & \multicolumn{1}{c}{testA} & \multicolumn{1}{c|}{testB} & \multicolumn{1}{c}{val} & \multicolumn{1}{c}{testA} & \multicolumn{1}{c|}{testB} & val(U) & test(U) & \multicolumn{1}{c||}{test(F)} &static &motion &static &motion & & \\\midrule

CRIS & $\text{\red{\xmark}}$ &66.68 &70.62 &59.93 &56.94 &64.20 &46.97 &55.91 &57.05 &58.62 &68.66 &52.53 &65.67 &45.98 &46.21 &47.62 \\
\citep{wang2022cris} & $\text{\green{\cmark}}$ &67.61 &71.89 &61.81 &57.26 &65.04 &47.25 &58.05 &58.93 &60.97 
 &70.14 &54.25 &67.13 &49.05 &47.25 &49.43 \\
& \textbf{Diff} & \greencell{+0.93} & \greencell{+1.27} & \greencell{+1.88} & \graycell{+0.32} & \greencell{+0.84} & \graycell{+0.28} & \greencell{+2.14} & \greencell{+1.88} & \greencell{+2.35} & \greencell{+1.48} & \greencell{+1.94} & \greencell{+1.46} & \greencell{+3.07} & \greencell{+1.04} & \greencell{+1.81} \\ \midrule

LAVT & $\text{\red{\xmark}}$ &72.72 &75.74 &68.56 &61.86 &67.97 &54.30 &61.24 &62.09 &64.48 &71.52 &53.28 &70.84 &46.58 &52.28 &57.48 \\
\citep{yang2022lavt} & $\text{\green{\cmark}}$ &73.07 &76.13 &69.41 &63.05 &67.95 &54.36 &63.93 &64.12 &66.24 &72.85 &56.21 &71.41 &50.11 &53.32 &59.17 \\
& \textbf{Diff}  & \graycell{+0.35} & \graycell{+0.39} & \greencell{+0.85} & \greencell{+1.19} & \graycell{+0.28} & \graycell{+0.06} & \greencell{+2.69} & \greencell{+2.03} & \greencell{+1.76} & \greencell{+1.33} & \greencell{+2.93} & \greencell{+0.57} & \greencell{+3.53} & \greencell{+1.04} & \greencell{+1.69} \\ \midrule

DMMI & $\text{\red{\xmark}}$ &72.85 &75.53 &69.53 &63.18 &67.98 &55.53 &61.75 &62.54 &66.94 &73.41 &51.62 &71.54 &45.83 &52.89 &55.94\\
\citep{hu2023beyond} & $\text{\green{\cmark}}$ &73.31 &76.26 &69.96 &63.51 &68.66 &56.15 &63.24 &64.98 &68.26 &74.68 &53.72 &72.73 &47.42 &54.16 &57.45 \\
& \textbf{Diff} & \graycell{+0.46} & \greencell{+0.73} & \graycell{+0.43} & \graycell{+0.33} & \greencell{+0.68} & \greencell{+0.62} & \greencell{+1.49} & \greencell{+2.44} & \greencell{+1.32} & \greencell{+1.27} & \greencell{+2.10} & \greencell{+1.19} & \greencell{+1.59} & \greencell{+1.27} & \greencell{+1.51} \\ \midrule


ASDA & $\text{\red{\xmark}}$ &75.08 &77.58 &71.21 &66.97 &71.81 &57.92 &65.71 &65.38 &57.94 
&76.26 &54.08 &75.19 &48.03 &52.92 &37.29 \\ 
\citep{yue2024adaptive} & $\text{\green{\cmark}}$ &75.82 &78.15 &72.17 &67.98 &72.89 &58.79 &66.98 &67.26 &58.64  &76.79 &58.95 &76.25 &53.15 &54.17 &38.85 \\
& \textbf{Diff}  & \greencell{+0.74} & \greencell{+0.57} & \greencell{+0.96} & \greencell{+1.01} & \greencell{+1.08} & \greencell{+0.87} & \greencell{+1.27} & \greencell{+1.88} & \greencell{+0.70} & \graycell{+0.53} & \greencell{+4.07} & \greencell{+1.06} & \greencell{+5.12} & \greencell{+1.25} & \greencell{+1.56} \\
\midrule

DETRIS & $\text{\red{\xmark}}$ &74.06 &77.15 &71.08 &64.62 &71.86 &56.35 &64.29 &64.98 &67.36 
&75.87 &62.01 &75.13 &56.27 &58.36 &61.22 \\ 
\citep{huang2025densely} & $\text{\green{\cmark}}$ &74.94 &77.83 &71.90 &65.25 &72.09 &56.88 &64.94 &65.61 &68.35
&77.53 &65.23 &76.42 &58.43 &59.69 &63.03 \\
& \textbf{Diff}  & \greencell{+0.88} & \greencell{+0.68} & \greencell{+0.82} & \greencell{+0.63} & \graycell{+0.23} & \graycell{+0.53} & \greencell{+0.65} & \greencell{+0.63} & \greencell{+0.99} &
\greencell{+1.66} & \greencell{+3.22} & \greencell{+1.29} & \greencell{+2.16} & \greencell{+1.33} & \greencell{+1.81} \\
\bottomrule
\end{tabular}%
}
\vspace{-0.2cm}
\caption{\textbf{Overall comparison of the RIS models}. {\setlength{\fboxsep}{1pt}\colorbox{green!10}{Green}} cells indicate statistically significant improvement, and {\setlength{\fboxsep}{1pt}\colorbox{gray!10}{gray}} cells mean neutral. 
({\setlength{\fboxsep}{1pt}\colorbox{red!10}{Red}} cells mean significant performance drop, but we have no such case.)
}
\label{tab:overall_comparison}
\vspace{-0.3cm}
\end{table*}

\noindent
\textbf{Evaluation Metrics.}
We adopt three standard metrics for RIS: mIoU, oIoU, and Prec@$p$.
The mIoU is the average IoU values computed per image, while oIoU measures the ratio of total intersection to total union across all test objects (samples). Since oIoU tends to favor larger objects, we use mIoU as a complementary size-balanced measure. Prec@$p$ with $p \in \{0.5, 0.7, 0.9\}$ indicates the percentage of samples with IoU above threshold $p$.

\noindent
\textbf{Implementation Details.}
We evaluated five state-of-the-art baselines with and without our method: LAVT~\citep{yang2022lavt}, CRIS~\citep{wang2022cris}, DMMI~\citep{hu2023beyond}, ASDA~\citep{yue2024adaptive}, and DETRIS~\citep{huang2025densely}.
These models represent different fusion strategies: LAVT and DMMI employ early fusion, CRIS and ASDA utilize late fusion, while DETRIS adopts multi-level fusion.

We keep original architectures and hyperparameters intact, adding only a single linear projection layer to produce multimodal embeddings for our MRaCL objective.
For MRaCL-specific hyperparameters, we set $\alpha=0.1$, margin $m=12$, and temperature $\tau=0.07$ across all datasets, empirically determined through ablation in \cref{sec:exp:ablation}.
All baselines are reproduced on NVIDIA RTX A6000 GPUs for fair comparison. Additional results on the recent LatentVG~\citep{yu2025latent} are provided in \cref{appendix:ablation_latentVG}.

\vspace{-0.2cm}

\subsection{Results and Analysis}
\label{sec:results}

\noindent
\textbf{Overall Comparison on Existing Benchmarks.}
The four leftmost column groups in \cref{tab:overall_comparison} compares the average oIoU of competing RIS models with and without MRaCL loss. For all datasets and models, MRaCL consistently improves performance, demonstrating its general effectiveness.

An interesting observation is that the degree of improvement correlates with the ratio of motion-centric queries per dataset.
On G-Ref and Ref-ZOM, which are relatively rich in verb-centric phrases, our method achieves substantial gains, boosting the baseline models by 1.71 and 1.42 oIoU points on average.
Notably, this trend holds even on verb-sparse datasets; on RefCOCO and RefCOCO+, adding MRaCL loss still provides consistent improvements with an average gain of 0.78 and 0.59 oIoU points, respectively.

This implies models trained with MRaCL better capture the nuances of verb-driven relationships and their visual grounding. Also, our method enhances general multi-modal alignment beyond motion understanding, benefiting RIS performance across diverse query types.

\noindent
\textbf{Evaluation on Action-centric Benchmarks.}
The rightmost 3 columns in \cref{tab:overall_comparison} compares the RIS performance on our motion-centric benchmarks: M-Ref and M-Bench.
Results on M-Ref demonstrate that our method improves performance on both static and motion splits, with substantially larger gains on motion-centric queries.
This confirms that our approach is particularly effective when objects must be disambiguated through motion-related cues rather than static attributes describing their appearance.

Lastly, recall that MBench features highly ambiguous scenarios, with multiple similar objects performing diverse actions.
The evaluation results on MBench in the rightmost column of \cref{tab:overall_comparison} also show consistent improvements across all models with our method.
Even in this challenging scenario, our method leads to significant gains, showing its effectiveness in helping RIS models focus on motion expressions.

\begin{figure}[t]
    \centering
    \includegraphics[width=\linewidth]{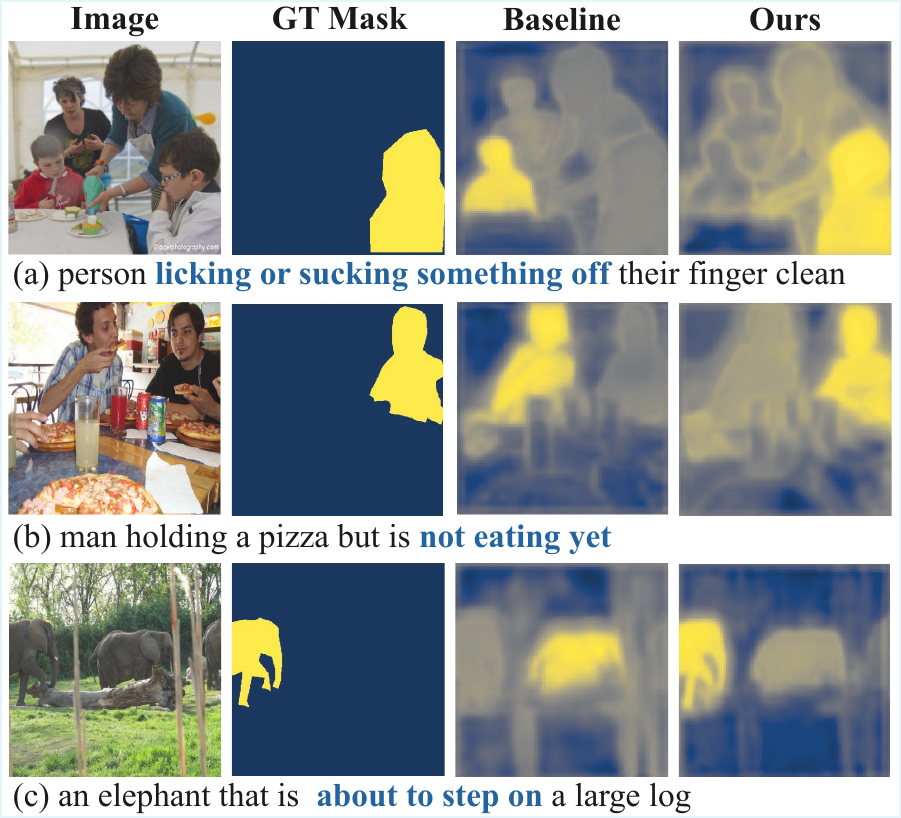}
    \vspace{-0.4cm}
    \caption{Visualization of activation maps with and without MRaCL on CRIS.} 
    \label{fig:actmap}
\end{figure}

\begin{figure}[t]
    \centering
    \includegraphics[width=\linewidth]{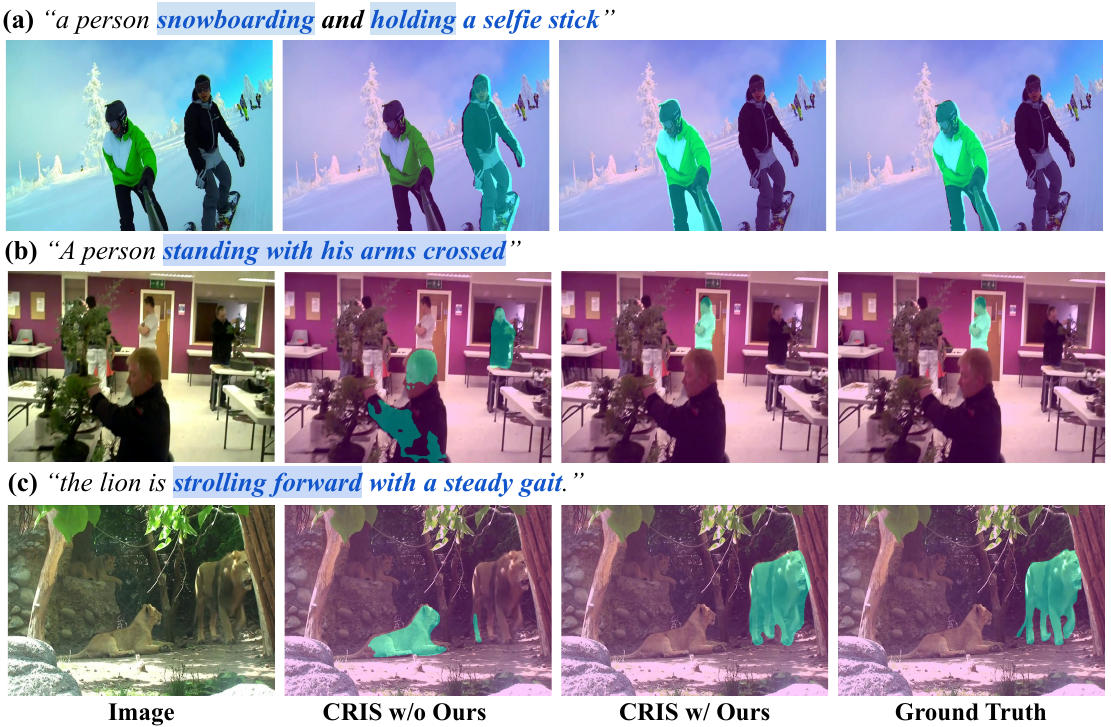}
    \caption{Qualitative results from MBench test set.} 
    \label{fig:qual}
    \vspace{-0.2cm}
\end{figure}

\noindent
\textbf{Qualitative Analysis.}
\cref{fig:qual,fig:actmap} both illustrate the effect of our method. We observe two consistent patterns of improvement. 
First, our method strengthens \textbf{\textit{action-level disambiguation}}. As seen in \cref{fig:qual}(a,c), CRIS with MRaCL correctly segments the target among multiple instances of the same category by grounding fine-grained action cues, where the baseline conflates visually similar subjects. The activation maps in \cref{fig:actmap}(a) further confirm this, as our method produces sharply focused activations on the acting person rather than diffusing across co-occurring instances. 
Second, MRaCL better captures \textbf{\textit{nuanced motion expressions}} that go beyond simple verb matching. In \cref{fig:actmap}(b,c), queries involving negation (``not eating yet'') and prospective intent (``about to step on'') require a deeper understanding of the described action context. Our method yields well-delineated activations aligned with the correct target in both cases, whereas the baseline fails to distinguish these subtle cues.
See App.~\ref{appendix:qual} for additional examples.


\begin{table}[t]
\centering
\setlength{\tabcolsep}{2pt}
\footnotesize
\resizebox{\columnwidth}{!}{%
\begin{tabular}{lcccccc}
\toprule
 & & & \multicolumn{3}{c}{\textbf{Prec@}} \\
\textbf{Method} & \textbf{mIoU} & \textbf{oIoU} & \textbf{0.5} & \textbf{0.7} & \textbf{0.9} \\
\midrule
CRIS & 59.17 & 55.91 & 67.95 & 55.45 & 15.27 \\
CRIS + L2 & 58.99 & 55.17 & 67.85 & 55.09 & 14.87 \\
CRIS + SimCSE & 59.77 & 56.09 & 68.15 & 55.45 & 14.42 \\
CRIS + InfoNCE (MCC) & 59.46 & 55.75 & 67.87 & 55.35 & 14.95 \\
CRIS + MRaCL (ours) & \textbf{60.87} & \textbf{58.05} & \textbf{69.73} & \textbf{57.45} & \textbf{15.85} \\
\bottomrule
\end{tabular}}
\vspace{-0.2cm}
\caption{\textbf{Comparison with various contrastive objectives},
\joonseok{measured on G-Ref validation set (UMD).}
}
\label{tab:contrastive_loss_comparison}
\end{table}

\begin{table}[t]
\centering
\footnotesize
\label{tab:semi_hard}
\setlength{\tabcolsep}{6pt}
\begin{tabular}{cccccc}
\toprule
\textbf{Ratio} & \textbf{mIoU} & \textbf{oIoU} & \textbf{P@0.5} & \textbf{P@0.7} & \textbf{P@0.9} \\
\midrule
0 & \textbf{60.87} & \textbf{58.05} & \textbf{69.73} & \textbf{57.45} & \textbf{15.85} \\
0.03 & 59.22 & 56.51 & 67.95 & 55.03 & 14.89 \\
0.1 & 46.84 & 42.42 & 51.35 & 44.76 & 10.03 \\
0.2 & 43.94 & 41.25 & 47.39 & 42.08 & 10.34 \\
\bottomrule
\end{tabular}
\vspace{-0.2cm}
\caption{\textbf{Ablation on semi-hard negative mining.} A ratio of 0 corresponds to our default augmentation-based alignment without semi-hard negatives.}
\end{table}


\noindent
\textbf{Comparison with other Contrastive Losses.}
In \cref{tab:contrastive_loss_comparison}, we compare various contrastive loss objectives on the G-Ref UMD validation set.
Starting from vanilla CRIS trained with cross-entropy only, we evaluate three contrastive alternatives: L2-based (Euclidean), SimCSE (cosine-based)~\citep{gao2021simcse}, and InfoNCE-based Meaning Consistency Constraint (MCC)~\citep{oord2018representation, nguyen2025vision}, alongside our MRaCL.

While L2-based loss slightly degrades performance and SimCSE and InfoNCE (MCC) yield marginal gains, MRaCL substantially outperforms all alternatives across all metrics.
We attribute this to the following factors: 1) distance-based losses such as L2 allow flexible representation scaling but lack the angular sensitivity needed for fine-grained multimodal alignment;
and 2) cosine similarity, used in both SimCSE and MCC, is prone to gradient saturation and anisotropic clustering~\citep{chen2024revisiting}, as further discussed in \cref{sec:method:MCL}.
In contrast, MRaCL leverages angular distance with explicit margin penalties, producing more discriminative embeddings that better capture subtle multimodal semantics.

\begin{table}[t]
\centering
\setlength{\tabcolsep}{2pt}
\footnotesize
\resizebox{\linewidth}{!}{
\begin{tabular}{ccc|cc}
\toprule
\textbf{Augment} & \textbf{MRaCL} & 
\textbf{Filtering} & \textbf{mIoU (\%)} & \textbf{oIoU (\%)} \\
\midrule
$\text{\red{\xmark}}$ & $\text{\red{\xmark}}$ & $\text{\red{\xmark}}$ & 59.28 & 55.91 \\
$\text{\red{\xmark}}$ & $\text{\green{\cmark}}$ & $\text{\red{\xmark}}$ & 59.45 & 56.02 \\
$\text{\green{\cmark}}$ & $\text{\red{\xmark}}$ & $\text{\red{\xmark}}$ & 59.31 & 57.34 \\
$\text{\red{\xmark}}$ & $\text{\green{\cmark}}$ & $\text{\green{\cmark}}$ & 59.89 & 56.11 \\
$\text{\green{\cmark}}$ & $\text{\green{\cmark}}$ & $\text{\red{\xmark}}$ & 60.18 & 57.33 \\
$\text{\green{\cmark}}$ & $\text{\green{\cmark}}$ & $\text{\green{\cmark}}$ & \textbf{60.87} & \textbf{58.05} \\
\bottomrule
\end{tabular}}
\vspace{-0.2cm}
\caption{\textbf{Ablation on proposed components on G-Ref.} Motion-centric augmented phrases (\cref{sec:method:hpconstruct}), angular distance-based MRaCL loss (\cref{sec:method:mracl_loss}), and filtering for potential false negatives (\cref{sec:method:MCL}).}
\label{tab:ablation}
\vspace{-0.2cm}
\end{table}

\noindent\textbf{Positive Alignment vs.\ Negative Mining.}
We examine whether semi-hard negative mining could further improve performance, considering negatives of two types: a different object performing the same action, or the same object performing a contradictory action. To approximate these, we replace the verb in each ground-truth caption with a randomly selected one, and train CRIS~\citep{wang2022cris} on G-Ref (UMD) with varying mixing ratios of such negatives.

As shown in \cref{tab:semi_hard}, incorporating semi-hard negatives consistently degrades performance, even at a ratio as small as 0.03.
We attribute this to the current state of RIS models: hard negative learning is generally effective when a model already performs coarsely and needs finer discrimination. 
However, as established in \cref{sec:intro} (\cref{tab:intro_motion_stat}), existing RIS models largely ignore motion cues and rely predominantly on static appearance.
Since these models have not yet learned to leverage motion semantics at all, hard negatives can act as noise rather than useful training signal.
This motivates our positive alignment strategy, which first provides a clear directional signal toward motion understanding as a necessary foundation before any discriminative negative objective can be effective.

\vspace{-0.3cm} \noindent
\subsection{Ablation Study}
\label{sec:exp:ablation}

\vspace{-0.1cm} \noindent
\noindent\textbf{Effectiveness of Components.}
We analyze the contribution of each component in our framework using CRIS~\citep{wang2022cris} on G-Ref (UMD) validation set.
Specifically, we compare combinations of three design factors: 1) using action-centric phrases as additional contrastive anchors ({Augmentation}; \cref{sec:method:hpconstruct}), 2) applying our MRaCL loss based on angular distance ({MRaCL}; \cref{sec:method:mracl_loss}), and 3) similarity-based filtering to eliminate potential false negatives ({Filtering}; \cref{sec:method:MCL}).

\begin{table}[t]
\centering
\footnotesize
\setlength{\tabcolsep}{3pt}
\resizebox{\linewidth}{!}{
\begin{tabular}{ll|ccccc}
\toprule
\multicolumn{2}{c|}{\textbf{Hyperparameter}} & \multicolumn{2}{c}{\textbf{val(UMD)}} & \multicolumn{3}{c}{\textbf{Prec(val)}} \\
& & \textbf{mIoU} & \textbf{oIoU} & \textbf{0.5} & \textbf{0.7} & \textbf{0.9} \\
\midrule
\multirow{6}{*}{Margin ($m$)} & 4 & 59.13 & 56.04 & 68.05 & 54.82 & 15.01 \\ 
& 8 & 59.66 & 56.80 & 68.71 & 56.85 & 15.08 \\
& 10 & 60.08 & 57.51 & 69.36 & 57.16 & 15.12 \\
& \textbf{12} & \textbf{60.87} & \textbf{58.05} & \textbf{69.73} & \textbf{57.45} & \textbf{15.85} \\
& 15 & 60.06 & 56.83 & 69.04 & 57.09 & 15.22 \\
& 20 & 59.26 & 55.88 & 67.97 & 55.56 & 15.24 \\
\midrule
& 0.35 & 59.54 & 55.93 & 68.16 & 55.29 & 15.51 \\
& 0.40 & 60.04 & 56.58 & 69.02 & 56.48 & 15.57 \\ 
Filtering & 0.45 & 60.11 & 57.02 & 69.28 & 56.84 & 15.42 \\
Threshold ($\nu$) & \textbf{0.50} & \textbf{60.87} & \textbf{58.05} & \textbf{69.73} & \textbf{57.45} & \textbf{15.85} \\ 
& 0.60 & 58.68 & 55.02 & 67.91 & 55.33 & 14.85 \\
& 0.70 & 59.52 & 56.12 & 68.01 & 56.35 & 15.62 \\
\midrule
& 0.05 & 60.17 & 57.25 & 69.12 & 56.33 & 15.93 \\ 
Metric & \textbf{0.10} & \textbf{60.87} & \textbf{58.05} & \textbf{69.73} & \textbf{57.45} & \textbf{15.85} \\ 
Loss & 0.15 & 60.13 & 57.68 & 68.63 & 57.38 & 15.53 \\ 
Weight ($\alpha$) & 0.10 & 59.64 & 56.13 & 68.51 & 55.47 & 15.23 \\ 
& 0.25 & 59.43 & 55.54 & 68.35 & 55.13 & 14.01 \\ 
\bottomrule
\end{tabular}}
\vspace{-0.2cm}
\caption{\textbf{Ablation on hyperparameter selection.} Experiments were conducted on G-Ref.}
\label{tab:refcocog_abl}
\vspace{-0.2cm}
\end{table}

In \cref{tab:ablation}, applying MRaCL alone (row 2) leads to stronger performance upon the baseline (row 1), verifying our hypothesis in \cref{sec:method:mracl_loss}.
Adding false negative filtering (row 4) further boosts the performance, particularly in oIoU.
Augmenting motion-centric expressions alongside original captions (row 5) further boosts the performance, guiding the model to better align with actions.
Applying all ideas at the same time (row 6) completes our proposed method, achieving the strongest performance.

\noindent\textbf{Ambiguity Filtering Strategy.}
Our data augmentation strategy exposes the model to more action-centric expressions by extracting verb phrases from original captions. However, 
ambiguity arises when multiple objects in the same category perform similar actions within a single image. To mitigate this, we apply a pre-filtering strategy that retains only training samples where the target's category appears once. \cref{tab:ambiguity_filtering} validates this choice: the application of stricter filtering thresholds consistently improves performance, and the strictest setting (Exclude $> 1$) achieves the best results in both validation and test sets. This confirms that reducing intra-class ambiguity produces a cleaner training signal for learning motion semantics.

\subsection{Effects of Hyperparameters}
\label{sec:exp:hyperparams}

\noindent
\textbf{Margin ($m$).}
\cref{tab:refcocog_abl} reports performance under varying margin values with fixed $\alpha = 0.1$ and $\nu = 0.5$.
A margin that is too small fails to enforce sufficient separation between positive and negative samples, while an excessively large margin causes representation collapse and degrades performance.
We find that a moderate margin of $m=12$ achieves the best results, providing an effective balance for contrastive alignment.

\begin{table}[t]
\centering
\setlength{\tabcolsep}{6pt}
\footnotesize
\begin{tabular}{lcccc}
\toprule
\textbf{Filtering Option} & \multicolumn{2}{c}{\textbf{mIoU}} & \multicolumn{2}{c}{\textbf{oIoU}} \\
\cmidrule(lr){2-3}\cmidrule(lr){4-5}
 & \textbf{val} & \textbf{test} & \textbf{val} & \textbf{test} \\
\midrule
No Filtering         & 58.96 & 57.98 & 56.71 & 57.66 \\
Exclude $> 3$        & 59.26 & 59.96 & 57.21 & 57.95 \\
Exclude $> 2$        & 60.54 & 60.38 & 57.83 & 58.39 \\
Exclude $> 1$ (Ours) & \textbf{60.87} & \textbf{60.47} & \textbf{58.05} & \textbf{58.95} \\
\bottomrule
\end{tabular}
\vspace{-0.2cm}
\caption{\textbf{Ablation on ambiguity filtering.}
The strictest threshold achieves the best results on G-Ref.}
\vspace{-0.2cm}
\label{tab:ambiguity_filtering}
\end{table}

\noindent
\textbf{Filtering Threshold ($\nu$).}
As discussed in \cref{sec:method:MCL}, we filter out potential false negatives during contrastive training. 
\cref{tab:refcocog_abl} shows that a moderate threshold around 0.5 results in consistent improvement, likely by eliminating noisy negatives that exhibit high similarity with the anchor.
Higher thresholds are less effective as fewer false negatives are detected, diminishing the benefit of filtering.
This trend is further supported by \cref{tab:ablation}, where the optimal threshold leads to clear gains when combined with verb-centric supervision (row 4 vs. row 5).

\noindent
\textbf{Metric Loss Weight ($\alpha$).}
We vary the metric loss weight $\alpha$, which determines the relative contribution of $\mathcal{L}_{\text{MRaCL}}$ compared to the segmentation loss $\mathcal{L}_{\text{seg}}$, with fixed $m = 12$ and $\tau = 0.07$.
As seen in \cref{tab:refcocog_abl}, $\alpha = 0.1$ achieves the best performance, where larger weights overemphasize the metric loss, slightly dropping the performance.
\vspace{-0.2cm} \noindent
\section{Conclusion}
\label{sec:conclusion}
\vspace{-0.2cm} \noindent
In this work, we address the limitation of existing Referring Image Segmentation (RIS) models in understanding motion-centric expressions. 
Specifically, we propose to leverage augmented action-centric queries to supplement 
training examples, and introduce \textbf{M}ultimodal \textbf{Ra}dial \textbf{C}ontrastive \textbf{L}oss (\textbf{MRaCL}), which performs contrastive learning on fused multimodal embeddings.
By utilizing angular distance as the similarity metric, MRaCL overcomes gradient saturation and anisotropy issues inherent in cosine-based methods.
Our method leads to superior performance across various RIS models and diverse benchmarks, with particularly significant improvements on test examples requiring a deeper understanding of motions.
These results highlight the effectiveness of our method in enhancing both linguistic and visual comprehension for RIS, providing a robust framework for tackling motion-oriented scenarios.
As a future work, a similar idea could be applied to the Referring Video Object Segmentation (RVOS) task, requiring even finer-grained understanding of actions.

\newpage
\section*{Acknowledgments}
This work was also supported by the SOFT Foundry Institute at SNU, Samsung Electronics, Youlchon Foundation, National Research Foundation of Korea (NRF) grants (RS-2021-NR05515, RS-2024-00336576, RS-2023-0022663, RS-2025-25399604, RS-2024-00333484), and the Institute for Information \& Communication Technology Planning \& Evaluation (IITP) grants (RS-2022-II220264, RS-2024-00353131) funded by the Korean government.

\bibliography{custom}

\section*{Checklist}



\begin{enumerate}

  \item For all models and algorithms presented, check if you include:
  \begin{enumerate}
    \item A clear description of the mathematical setting, assumptions, algorithm, and/or model. [\textbf{Yes}]
    \item An analysis of the properties and complexity (time, space, sample size) of any algorithm. [\textbf{No}]
    \item (Optional) Anonymized source code, with specification of all dependencies, including external libraries. [\textbf{Yes}] We will release the code if our paper is accepted.
  \end{enumerate}

  \item For any theoretical claim, check if you include:
  \begin{enumerate}
    \item Statements of the full set of assumptions of all theoretical results. [\textbf{Not Applicable}]
    \item Complete proofs of all theoretical results. [\textbf{Not Applicable}]
    \item Clear explanations of any assumptions. [\textbf{Not Applicable}]     
  \end{enumerate}

  \item For all figures and tables that present empirical results, check if you include:
  \begin{enumerate}
    \item The code, data, and instructions needed to reproduce the main experimental results (either in the supplemental material or as a URL). [\textbf{Yes}] Hyperparmeters for the main results are explained in ~\cref{sec:exp}, data preperation processes are explained in \cref{appendix:mref} and \cref{appendix:mbench_construction}. Code and data will be released on public repository if the paper is accepted. 
    \item All the training details (e.g., data splits, hyperparameters, how they were chosen). [\textbf{Yes}] 
    \item A clear definition of the specific measure or statistics and error bars (e.g., with respect to the random seed after running experiments multiple times). [\textbf{Yes}]
    \item A description of the computing infrastructure used. (e.g., type of GPUs, internal cluster, or cloud provider). [\textbf{Yes}]. Details are explained in ~\cref{sec:exp:setting}.
  \end{enumerate}

  \item If you are using existing assets (e.g., code, data, models) or curating/releasing new assets, check if you include:
  \begin{enumerate}
    \item Citations of the creator If your work uses existing assets. [\textbf{Yes}]. Our method was implemented upon the baselines in \cref{sec:related} and \cref{sec:exp:setting}.
    \item The license information of the assets, if applicable. [\textbf{Not Applicable}]
    \item New assets either in the supplemental material or as a URL, if applicable. [\textbf{Yes}]. We plan to provide open access to our MBench evaluation dataset via a public repository if paper is accepted.
    \item Information about consent from data providers/curators. [\textbf{Not Applicable}]
    \item Discussion of sensible content if applicable, e.g., personally identifiable information or offensive content. [\textbf{No}]
  \end{enumerate}

  \item If you used crowdsourcing or conducted research with human subjects, check if you include:
  \begin{enumerate}
    \item The full text of instructions given to participants and screenshots. [\textbf{Not Applicable}]
    \item Descriptions of potential participant risks, with links to Institutional Review Board (IRB) approvals if applicable. [\textbf{Not Applicable}]
    \item The estimated hourly wage paid to participants and the total amount spent on participant compensation. [\textbf{Not Applicable}]
  \end{enumerate}

\end{enumerate}

\clearpage
\appendix
\thispagestyle{empty}
\onecolumn

\setcounter{page}{1}
\pagenumbering{roman}
\crefalias{section}{appendix}
\crefalias{subsection}{appendix}
\crefalias{subsubsection}{appendix}

\setcounter{table}{0}
\setcounter{figure}{0}
\setcounter{equation}{0}
\renewcommand{\thetable}{\Roman{table}}
\renewcommand{\thefigure}{\Roman{figure}}
\renewcommand{\theequation}{\roman{equation}}

\aistatstitle{Towards Motion-aware Referring Image Segmentation: \\
Supplementary Materials}

\section{M-Ref Motion \& Static Split Examples}
\label{appendix:mref}

\begin{figure*}[h]
    \centering
    \includegraphics[width=\linewidth]{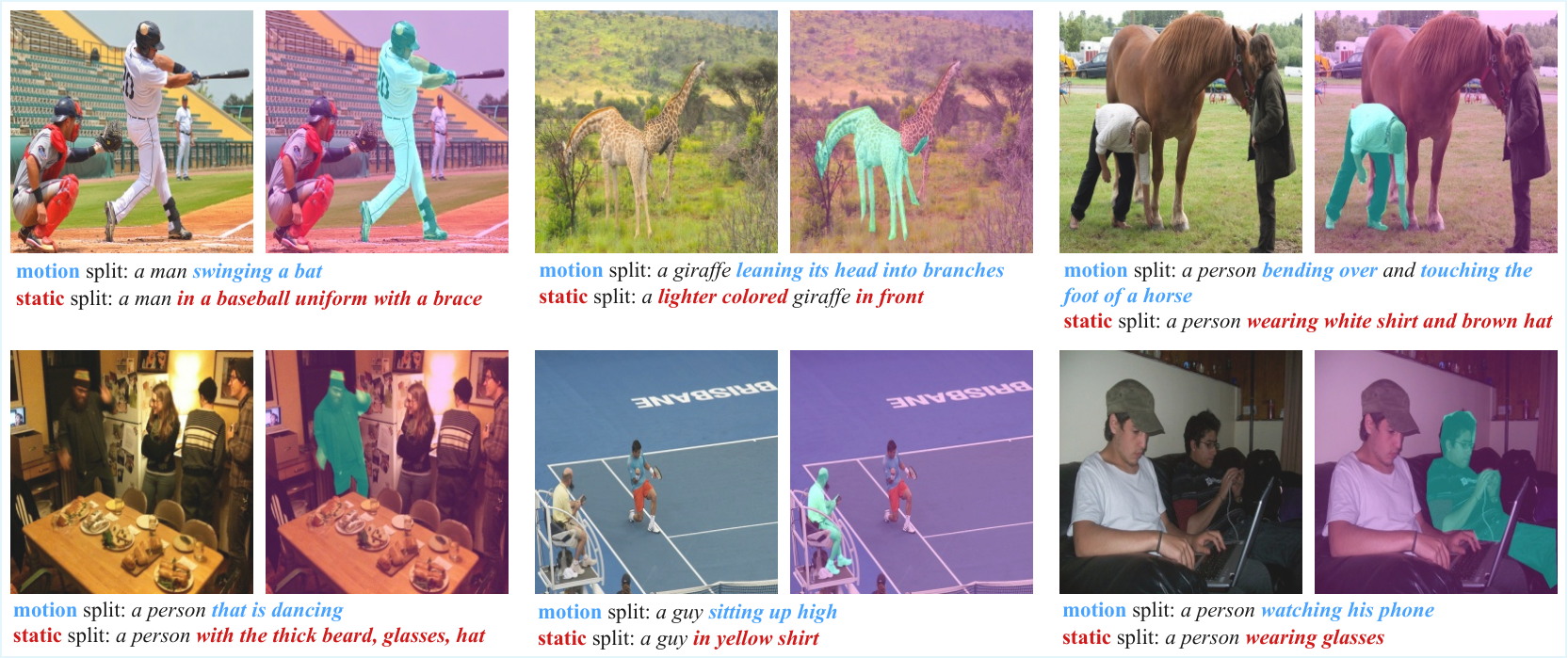}
    \caption{Examples of motion and static descriptions in M-Ref.}
    \label{fig:mref}
\end{figure*}

We construct a manually annotated split of the G-Ref UMD test set by categorizing each referring expression as either static or motion-centric.
Static queries primarily describe appearance-related or position-based attributes—such as clothing color, physical traits, or spatial relations (\textit{e.g.}, ``a man in a baseball uniform'', ``a person with the thick beard, glasses and a hat'')—which leverage visually stable cues and require minimal reasoning about motion.
In contrast, motion-centric queries emphasize the dynamic behavior or transient actions of the target object, including expressions like ``swinging a bat'', ``bending over and touching the foot of a horse'', or ``holding a tennis racket while balancing a ball''.
While both types of expressions may refer to the same visual target, motion-oriented queries provide more explicit cues about the object’s ongoing action, demanding finer understanding and contextual interpretation from the model.
This curated split reveals the inherent linguistic and semantic gap between static and dynamic references, serving as a diagnostic benchmark to evaluate how well RIS models can handle both appearance-based and motion-aware expressions.
\cref{fig:mref} illustrates a few examples of image captions in both motion and static splits.

\begin{figure}
    \centering
    \includegraphics[width=\linewidth]{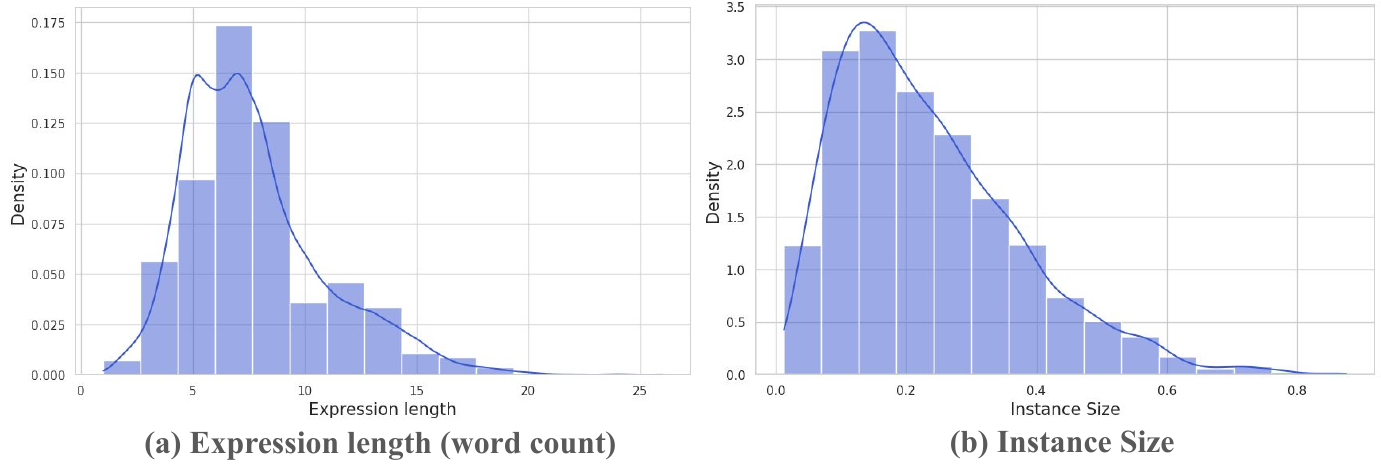}
    \caption{Analysis of expression length and instance size} 
    \label{fig:expression_length_instance_size}
\end{figure}

\begin{figure}
    \centering
    \includegraphics[width=\linewidth]{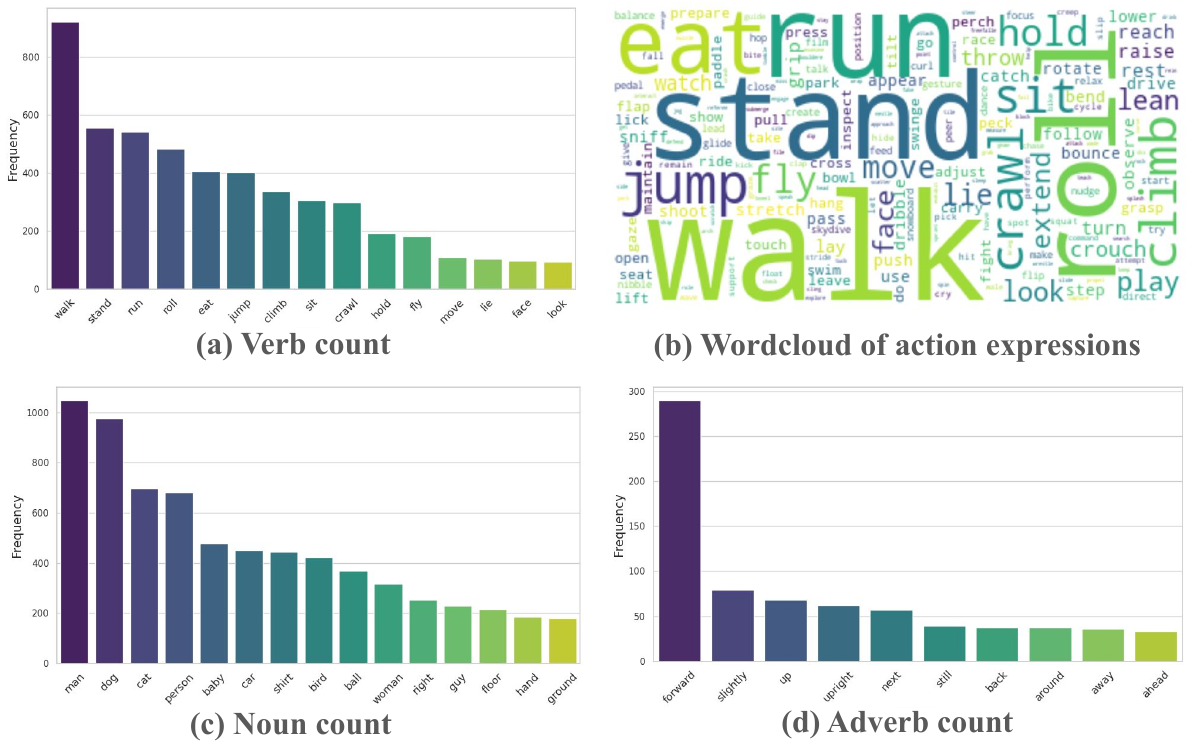}
    \caption{Most frequent verbs, nouns and adverbs} 
    \label{fig:word_counts}
\end{figure}

\begin{figure*}
    \centering
    \includegraphics[width=\linewidth]{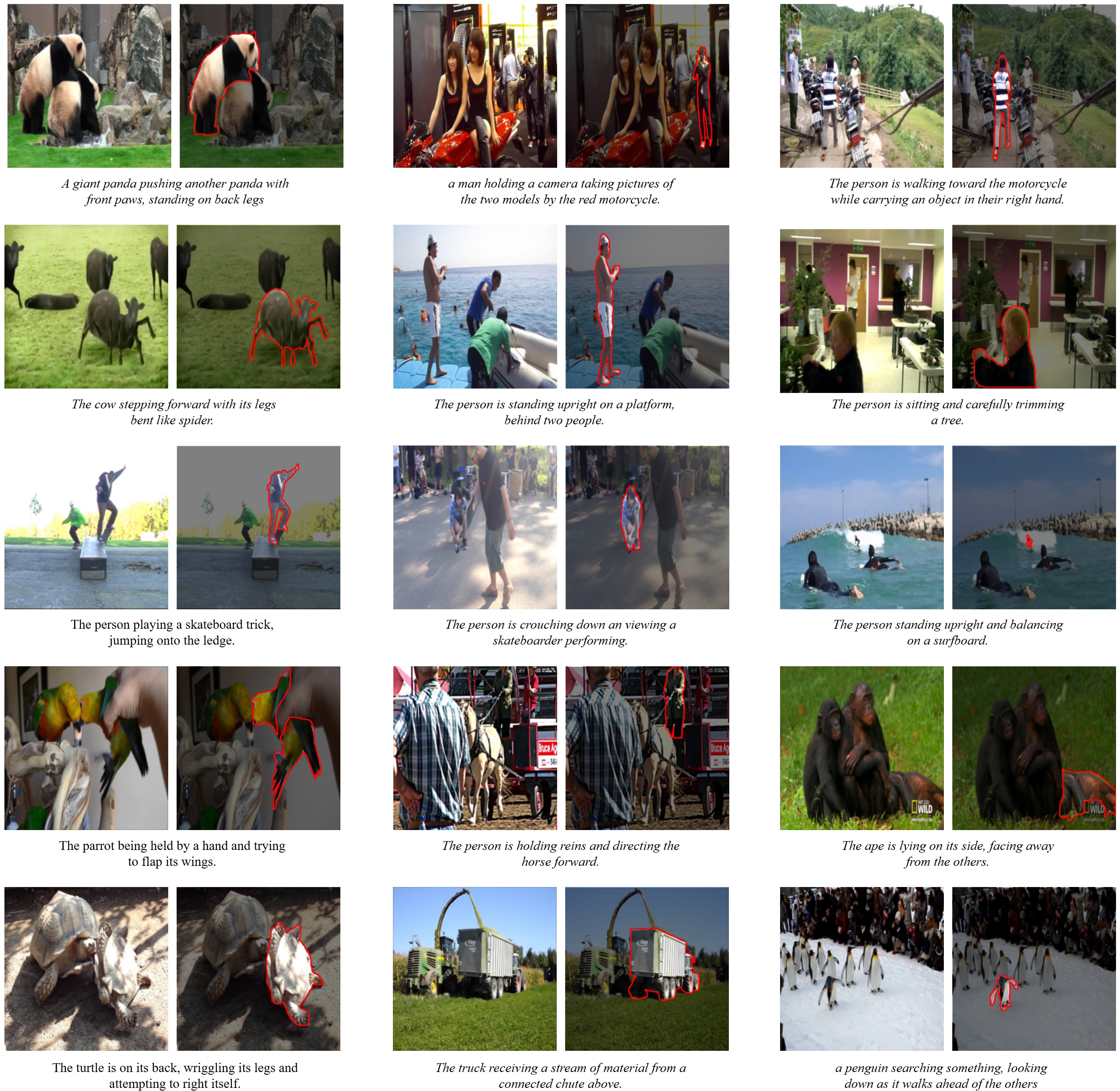}
    \caption{Visualization of MBench image and its highlighted mask}
    \label{fig:mbench_image}
\end{figure*}

\section{MBench Dataset Details}
\label{appendix:mbench}
Our MBench consists of image-text pairs where the target instance is differentiated from other objects of the same category using an action-centric expression.
It contains 8,200 instance annotations with referring expressions of varying lengths. Each expression contains 7.72 words on average and up to 26 words.
Following~\citep{chen2024revisiting}, we compute the normalized instance size relative to the image size as
\begin{equation}
    \sqrt{\frac{\text{instance area}}{\text{image area}}},
\end{equation}
and its distribution, illustrated in \cref{fig:expression_length_instance_size}, peaks at 0.137.
The distribution exhibits a long tail, with a maximum value of 0.875, indicating that our dataset targets instances of various sizes.
Analyzing the referring expressions using the \texttt{spaCy} library, we identify 1,268 unique lemmatized words.
The 15 most frequent verbs are \emph{walk, stand, run, roll, eat, jump, climb, sit, crawl, hold, fly, move, lie, face, look}, suggesting that our dataset focuses on casual, everyday activities.
We plot the frequencies of 15 most frequent verbs and nouns, and the 10 most frequent adverbs in \cref{fig:word_counts}.

In \cref{fig:mbench_image}, we illustrate a few examples of  the original images and their highlighted masks of MBench.
The top-left image shows two giant pandas doing different actions and the model is expected to focus on the visual aspect of `pushing' and `standing' and understand the relationship between pandas (\textit{e.g.}, `pushing another panda') to correctly find the left panda.
Action-centric expressions also include passive verbs such as `being held'.
Using these referring expressions, we aim to segment not only animate category objects (\textit{e.g.}, ‘person’, ‘cow’, ‘ape’) but also inanimate objects (\textit{e.g.}, ‘truck’).
Targeted instances are of various sizes, with smaller objects such as the third image in the right column (``The person standing upright and balancing on a surfboard'') of \cref{fig:mbench_image} being harder to find.
Instances may be partially occluded like the fourth image in the center column (``The person is holding reins and directing the horse forward'') and other objects belonging to the same category may be numerous such as the bottom right image (``a penguin searching something, looking down as it walks ahead of the others'') making the task especially difficult.

\section{Details on MBench Construction}
\label{appendix:mbench_construction}
Although recently developed RIS datasets contain complicated examples, most of them are still distinguished by appearance cues. 
In order to evaluate the performance of RIS models with respect to their understanding of action-centric cues in referring expressions, we develop a custom evaluation benchmark called \textbf{MBench}, shortened from `motion expression centric benchmark'.
MBench provides a targeted benchmark to assess the ability of a model trained on conventional RIS datasets on action-oriented expressions.
By focusing on descriptions where visual disambiguation stems from motion or behavior rather than appearance, MBench serves as a controlled benchmark for assessing the transferability of verb-centric grounding capabilities.

In order to leverage rich action-related cues, we build upon three text-grounded video segmentation datasets; namely, A2D-Sentences~\citep{gavrilyuk2018actor}, Ref-Youtube-VOS~\citep{seo2020urvos}, and ViCaS~\citep{athar2024vicas}.

\vspace{0.1cm} \noindent
\textbf{Frame Sampling and Category-based Filtering.}
We first divide each video into four equal-length segments, and sample one frame from each of them to ensure diversity.
This step is skipped for A2D, since each annotation already corresponds to a preselected frame.

Then, we annotate a referring expression for each object instance in the selected frames, describing its visual context and behavior within the scene.
Considering our goal to evaluate action-centric disambiguation, it is important that multiple co-occurring instances in the scene are distinguishable by their behavior. Thus, we keep frames where at least two or more instances belonging to the same category (\emph{e.g.}, woman, horse, cat) cannot be trivially differentiated by category name alone. 

If a frame contains only one instance of a given category, it is excluded from the benchmark. For this step, we use the category annotations if provided in the base set, where each annotated instance is labeled with a corresponding target category ID. Otherwise, we parse the main noun phrase from the referring expression as a pseudo-category. For example, from a caption like ``A man in shorts is exercising with a rubber band'', the parser derives candidate labels such as ``man in shorts'' and ``rubber band''. These extracted phrases serve as semantic object categories and are used for downstream filtering and evaluation.

We then determine whether each identified category belongs to a movable entity (\textit{e.g.}, human, animal, vehicle) based on a manually curated list.
In cases where pseudo-categories are used, we query a lightweight language model (Gemini~\citep{team2023gemini} 2.0 Flash Lite) with prompts like ``Is \{category\} capable of independent motion or typical actions?'' to assess movability.
Only objects confirmed as movable are retained.


\vspace{0.1cm} \noindent
\textbf{Caption Generation.}
For each frame that satisfies the criteria above, we proceed to the captioning stage. 
First, we overlay the  segmentation masks provided by each dataset as unfilled polygons on the image.
Following Yang et al.~\citep{yang2023set}, we additionally annotate an
object ID at the center of each instance to facilitate clear identification, as illustrated in \cref{fig:mbench_pipeline}.
This setup ensures that all instances within a frame are distinctly marked, allowing unambiguous reference and precise disambiguation during captioning.

Each marked frame is then fed into an LLM (\emph{e.g.}, we use GPT-4o) via a two-stage `visual chain-of-thought prompting' strategy.
In the first stage, we verify if the identified objects of the same category are performing distinct and recognizable actions (Movable Action Filtering).
If so, we proceed to the second stage, where we prompt the LLM to generate dense object-level captions focusing solely on the prominent action verb, excluding appearance, spatial relations, or ambiguous speculations.
We utilize multiple prompt templates to encourage diversity in expression.

\vspace{0.1cm} \noindent
\textbf{Text Refinement.}
We observe that the initially generated captions often contain noisy patterns, such as explicit instance IDs (\textit{e.g.}, ``the giant panda marked `2' ''), redundant action descriptions, or unnecessary appearance expressions.
To improve clarity and consistency of the captions, we rewrite them into concise action descriptions focused on individual objects, using a lightweight language model (Llama-3.1-8B~\citep{grattafiori2024llama}).
This step ensures action-oriented queries, aligned with our goal of producing this benchmark set.

Lastly, we manually review a subset of the resulting data to construct a high-quality validation set, discarding samples if their generated captions fail to reflect the object's actual behavior or if multiple instances exhibit indistinguishable actions.
Annotators are presented with the marked frame and the generated caption, and are asked to verify both the semantic accuracy and visual alignment of the descriptions.
Through this pipeline, we obtain 8,200 action-focused evaluation samples (instances) for the RIS task, requiring fine-grained object understanding in dynamic visual scenes.
We will publicly release this benchmark set upon acceptance.



\begin{figure*}[t]
\centering
\footnotesize
\includegraphics[width=0.8\linewidth]{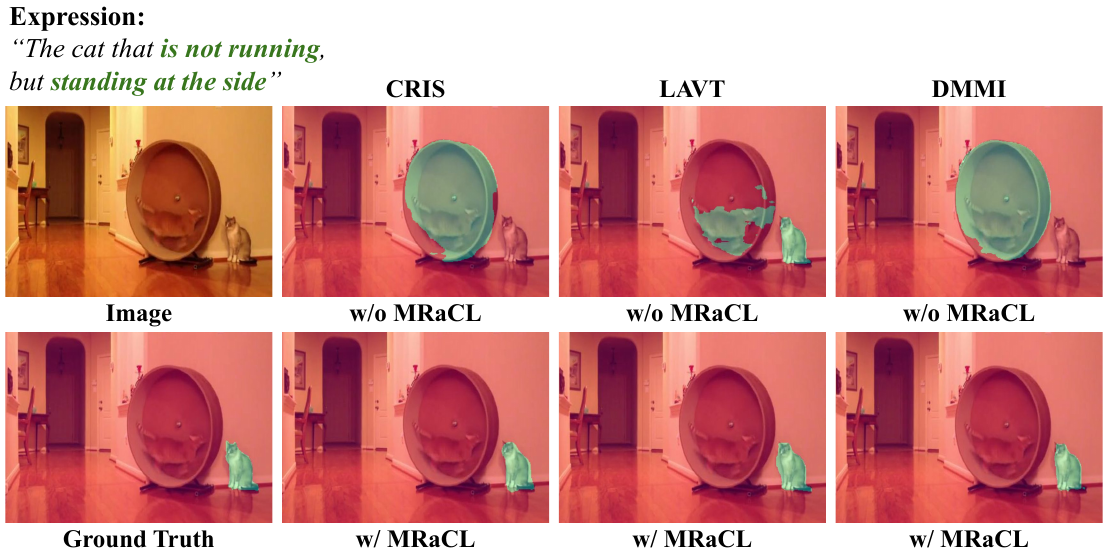}\\[0.3em]
\includegraphics[width=0.8\linewidth]{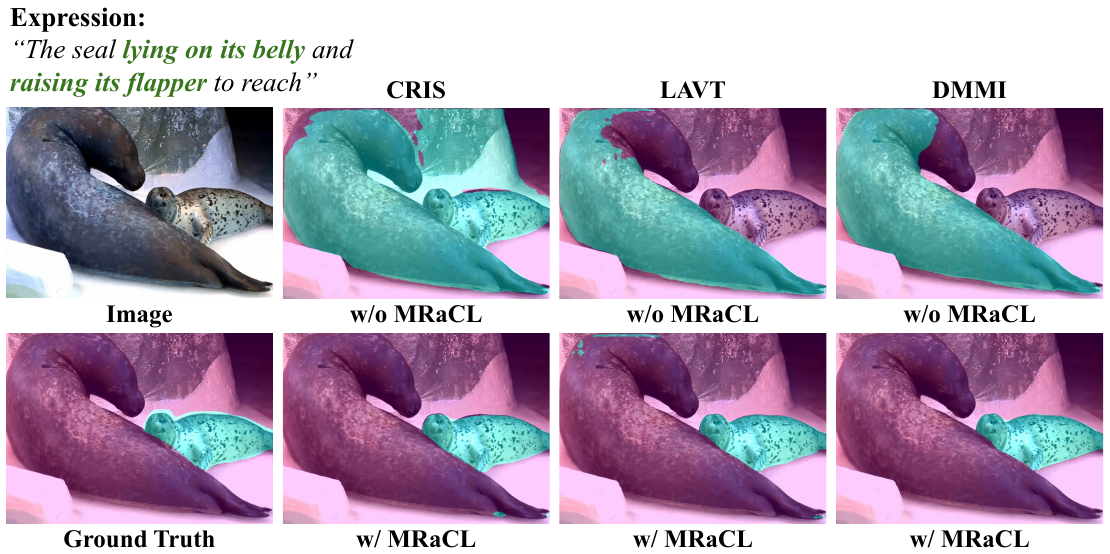}
\caption{\textbf{Additional visualizations of predictions from the MBench test set.} For each example, the left-most column presents the input expression, followed by the input image, and the ground truth mask overlaid on the image for reference. Subsequent columns across each row exhibit the comparative visualizations of model predictions before and after applying our MRaCL based framework.}
\label{fig:qualitative_comparison}
\end{figure*}

\section{Additional result on more advanced baseline }
\label{appendix:ablation_latentVG}

We additionally compare with a recently published state-of-the-art RIS model Latent-VG~\citep{yu2025latent}. As shown in the table below, we observe consistent performance improvements across all three datasets, aligning with the trends reported in \cref{tab:overall_comparison} (oIoU scores). For a fair comparison, we reproduce all baselines under the same setting as noted in~\cref{sec:exp:setting}. 

\begin{table}[h!]
\centering
\caption{\textbf{Additional results with LatentVG~\citep{yu2025latent} as baseline (oIoU).} Applying MRaCL to a more recent and stronger baseline yields consistent improvements across all splits of all three datasets.}
\label{tab:latentVG}
\setlength{\tabcolsep}{3pt}
\resizebox{0.75\linewidth}{!}{
\begin{tabular}{l|ccc|ccc|cc}
\toprule
\multirow{2}{*}{\textbf{Method}}
& \multicolumn{3}{c|}{\textbf{RefCOCO (UNC)}}
& \multicolumn{3}{c|}{\textbf{RefCOCO+ (UNC+)}}
& \multicolumn{2}{c}{\textbf{G-Ref (UMD)}} \\
\cmidrule(lr){2-4}\cmidrule(lr){5-7}\cmidrule(lr){8-9}
& \textbf{Val} & \textbf{TestA} & \textbf{TestB}
& \textbf{Val} & \textbf{TestA} & \textbf{TestB}
& \textbf{Val} & \textbf{Test} \\
\midrule
Latent-VG (reproduced)
& 76.23 & 78.39 & 73.33
& 69.16 & 72.83 & 62.18
& 69.07 & 69.83 \\
\textbf{Latent-VG + MRaCL}
& \textbf{76.94} & \textbf{78.87} & \textbf{73.95}
& \textbf{69.49} & \textbf{73.51} & \textbf{62.56}
& \textbf{70.02} & \textbf{70.66} \\
\bottomrule
\end{tabular}
}
\vspace{-0.3cm}
\end{table}

\section{More Qualitative Results}
\label{appendix:qual}

We present additional qualitative results of applying our framework to the state-of-the-art baselines, evaluated on the MBench dataset.
As shown in \cref{fig:qualitative_comparison}, models trained with our method demonstrate consistent improvements in segmenting target entities described by action-centric referring expressions.
These qualitative examples further illustrate how our proposed method enables RIS models to capture nuanced verb-centric semantics without requiring full model retraining, in line with the goals outlined in our approach.

For example, in the top row of \cref{fig:qualitative_comparison}, the expression ``\textit{The cat that is not running, but standing at the side}'' requires grounding a subtle, non-salient action.
Baselines often misidentify the subject, whereas those equipped with our method correctly localize the stationary cat by capturing fine-grained pose semantics.
In the second case, the expression ``\textit{The seal lying on its belly and raising its flapper to reach}'' involves two concurrent actions.
Without our framework, baseline models fail to fully segment the active region or confuse nearby objects.
In contrast, models trained with ours successfully align the dual motions—body posture and limb movement—demonstrating improved understanding of compositional, action-centric expressions.



\end{document}